\documentclass[10pt,twocolumn,letterpaper]{article}

\usepackage{wacv}
\usepackage{times}
\usepackage{epsfig}
\usepackage{graphicx}
\usepackage{amsmath}
\usepackage{amssymb}
\usepackage{colortbl}

\def\wacvPaperID{177} 

\wacvfinalcopy 

\ifwacvfinal
\def\assignedStartPage{9876} 
\fi

\ifwacvfinal
\usepackage[breaklinks=true,bookmarks=false]{hyperref}
\else
\usepackage[pagebackref=true,breaklinks=true,colorlinks,bookmarks=false]{hyperref}
\fi

\ifwacvfinal
\else
\fi

\begin{document}

\title{Alleviating Over-segmentation Errors by Detecting Action Boundaries}

\author{
Yuchi Ishikawa$^{1, 2}$, Seito Kasai$^{1, 2}$, Yoshimitsu Aoki$^2$, Hirokatsu Kataoka$^1$\\
\\$^1$National Institute of Advanced Industrial Science and Technology (AIST)\\
$^2$Keio University\\
{\tt\small\{yuchi.ishikawa, seito-kasai, hirokatsu.kataoka\}@aist.go.jp, aoki@elec.keio.ac.jp}
}

\maketitle
\begin{abstract}
We propose an effective framework for the temporal action segmentation task, namely an \textbf{Action Segment Refinement Framework (ASRF)}.
Our model architecture consists of a long-term feature extractor and two branches: the Action Segmentation Branch (ASB) and the Boundary Regression Branch (BRB).
The long-term feature extractor provides shared features for the two branches with a wide temporal receptive field.
The ASB classifies video frames with action classes, while the BRB regresses the action boundary probabilities.
The action boundaries predicted by the BRB refine the output from the ASB, which results in a significant performance improvement.
Our contributions are three-fold: (i) We propose a framework for temporal action segmentation, the ASRF,
which divides temporal action segmentation into frame-wise action classification and action boundary regression.
Our framework refines frame-level hypotheses of action classes using predicted action boundaries.
(ii) We propose a loss function for smoothing the transition of action probabilities, and analyze combinations of various loss functions for temporal action segmentation.
(iii) Our framework outperforms state-of-the-art methods on three challenging datasets, offering an improvement of up to 13.7\% in terms of segmental edit distance and up to 16.1\% in terms of segmental F1 score.
Our code will be publicly available soon\footnote{https://github.com/yiskw713/asrf}.
\end{abstract}

\vspace{-5pt}
\section{Introduction}
In recent years, with the exponential increase in the number of videos uploaded on the Internet, more attention is now being paid to video analysis. 
One of the most active topics in video analysis is video classification, the goal of which is to classify a trimmed video into a single action label~\cite{AggarwalCSUR2011}. The rise of sophisticated architectures (e.g. two-stream convnets~\cite{SimonyanNIPS2014,FeichtenhoferCVPR2016}, 3D CNNs~\cite{TranICCV2015,CarreiraCVPR2017,HaraCVPR2018,s3d,slowfast,x3d}, (2+1)D CNN~\cite{TranCVPR2018}) and large-scale video datasets such as Kinetics~\cite{KayarXiv2017,Carreiraarxiv2019} and Sports-1M~\cite{KarpathyCVPR2014} has greatly improved the performance of video classification.

However, natural videos are untrimmed and may contain multiple action instances.
Therefore, analysis of these videos demands the recognition of action sequences.
Motivated by this, we address the task of temporal action segmentation, which aims to capture and classify each action segment of an untrimmed video into an action category.
Action segmentation has various potential applications in robotics, surveillance and the analysis of human activities.

\begin{figure}[t!]
\begin{center}
\includegraphics[width=83mm]{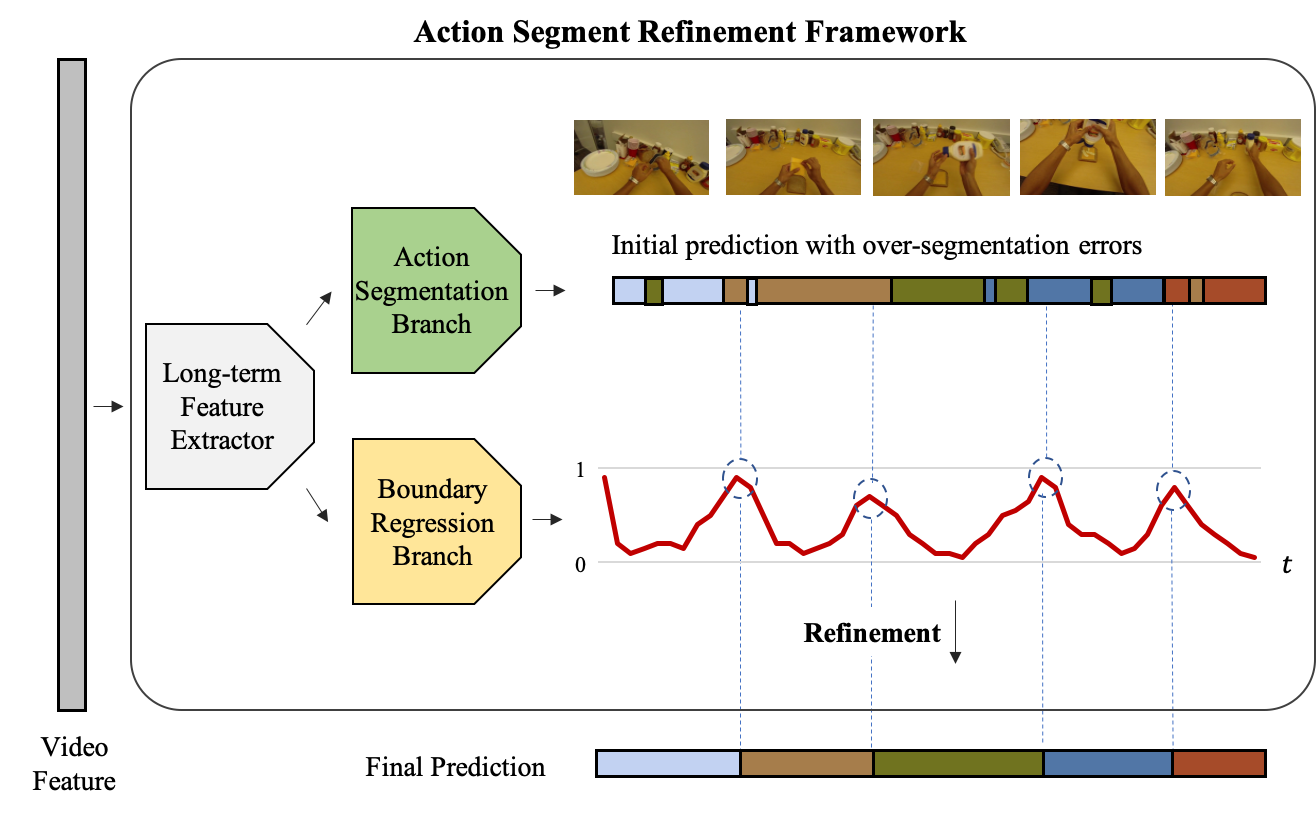}
\end{center}
\caption{
Overview of proposed framework. The Action Segment Refinement Framework (ASRF) consists of a long-term feature extractor(Section \ref{extractor}), an Action Segmentation Branch (ASB; Section \ref{asb}) and a Boundary Regression Branch (BRB; Section \ref{brb}). 
Video features are input for a long-term feature extractor, and it provides shared features for the ASB and the BRB.
Then these two branches output frame-wise action class predictions and boundary probabilities, respectively.
The ASRF reassigns frame-level action predictions from the ASB utilizing the action boundaries predicted by the BRB.
}
\label{fig:network}
\vspace{-5pt}
\end{figure}

Extant research typically approaches this task through two phases: i) extracting spatial or spatiotemporal features with 2D CNNs, two-stream convnets~\cite{SimonyanNIPS2014,FeichtenhoferCVPR2016} or 3D CNNs~\cite{CarreiraCVPR2017} and
ii) temporally classifying the extracted features using an RNN~\cite{bi-lstm} or temporal convolutional networks~\cite{tcn,deform,mstcn}.
These models claim better results on datasets with a small number of action classes and video clips than earlier approaches~\cite{slidingwindow1,slidingwindow2}.
However, it is difficult for these methods to recognize action segments, especially on large datasets with diverse action classes, which results in continuously fluctuating action class predictions, or \textit{over-segmentation errors}.

When considering temporal action segmentation, over-segmentation is a critical issue.
The prediction in the upper part of Figure~\ref{fig:network} shows an example prediction with over-segmentation errors.
This error is crucial when analyzing untrimmed videos with a sequence of actions.
For example, when analyzing the steps that are taken in cooking videos, over-segmentation will result in detecting extra steps.
Assuming that actions in videos exist as chunks and do not rapidly change, reducing these errors is essential for improving action segmentation performance.

To address this problem, we propose a framework for action segmentation that leverages predicted action boundaries for action segmentation, namely Action Segment Refinement Framework (ASRF).
The ASRF consists of a long-term feature extractor and two branches, an Action Segmentation Branch (ASB) and a Boundary Regression Branch (BRB) as in Figure \ref{fig:network}.
The long-term feature extractor expands the temporal receptive field and provides shared features for the following two branches. 
Then, the ASB broadly predicts frame-wise actions in a video, while the BRB detects action boundaries regardless of action class.
Our framework refines the outputs from the ASB using action boundaries predicted by the BRB.
The ASB and the BRB are complementary to each other, which enables the reduction of over-segmentation errors. In summary, the main contributions of this work are highlighted as follows:

\begin{enumerate}
\setlength{\itemsep}{0cm}
\setlength{\parskip}{0cm} 
\item We propose a simple but effective framework for action segmentation, an ASRF. Our framework decouples frame-wise action classification and action boundary regression, which enables the capturing of reliable segments and correct classification. Having a decoupled architecture enhances single model classification results by up to 10.6\% in terms of segmental F1 score and 10.8\% in terms of segmental edit distance. 
\item We propose a loss function for smoothing the transition of action probabilities and investigate appropriate combinations of loss functions for our framework. Combining this loss function with a class-weighted classification loss function enables a 8.2\% improvement of segmental F1 score and a 5.9\% improvement of segmental edit distance.
\item Our approach outperforms state-of-the-art methods on three challenging datasets for action segmentation: 50 Salads~\cite{50salads}, Georgia Tech Egocentric Activities (GTEA)~\cite{gtea} and the Breakfast dataset~\cite{breakfast}, by up to 16.1\% improvement of segmental F1 score, 13.7\% improvement of segmental edit distance and 2.6\% improvement for frame-wise accuracy.
\end{enumerate}


\section{Related Work}
\textbf{Video Representation.}
Action segmentation aims to classify each frame in a video into action classes, assuming that each frame contains a single action.
Extant methods are typically divided into two phases:
first extracting frame-wise spatio-temporal features by 2DCNNs\cite{tsm}, two-stream convnets~\cite{SimonyanNIPS2014,FeichtenhoferCVPR2016} or 3DCNNs~\cite{CarreiraCVPR2017,s3d,slowfast,channel-separated,x3d}, and then conducting the frame-wise classification.
Some works~\cite{ensemble_cnns,markov2,bi-lstm2,Mac_2019_ICCV} focus on extracting effective features for action segmentation.
However, our work focuses on the method for classifying action and feature extraction is beyond the scope of this work.
Following~\cite{mstcn}, we use I3D~\cite{CarreiraCVPR2017} features as input to our framework.

\textbf{Action Segmentation.}
Earlier approaches detect action segments using a sliding window and filter out redundant hypotheses with non-maximum suppression~\cite{slidingwindow1,slidingwindow2}.
Other approaches model the temporal action sequence with a Markov model~\cite{markov1,markov3} or an RNN~\cite{bi-lstm2} to classify frame-wise actions.
Lea \textit{et al.}~\cite{markov2} propose a spatiotemporal CNN for extracting features, which takes RGB and motion history images as input,
and uses a semi-Markovian model for jointly segmenting and classifying actions.

Following the success in the speech synthesis domain, some studies have adopted a temporal convolutional network from the WaveNet model~\cite{wavenet}.
Lea \textit{et al.}~\cite{tcn} propose two types of temporal convolutional networks (TCN) for action segmentation.
Lei \textit{et al.}~\cite{deform} propose a network with temporal deformable convolutions and a residual stream.
These approaches use temporal pooling, which help to capture long-range dependencies between actions.
However, pooling operations may lose temporal information that is indispensable for action segmentation.

To address this problem, Farha \textit{et al.}~\cite{mstcn} propose a multi-stage architecture, namely MS-TCN.
MS-TCN stacks several TCNs, which have dilated convolutions with residual connections.
They also propose a loss function for penalizing over-segmentation errors.
This architecture and the loss function enable the refinement of action segmentation results through each stage.
Zhang \textit{et al.}~\cite{bpgaussian} propose a bilinear pooling module and combines it with MS-TCN.
The above two methods are capable of capturing dependencies between actions and reducing over-segmentation errors.
Yifei \textit{et al.}~\cite{as_graph} propose Graph-based Temporal Reasoning Module, which can be added to top of action segmenation models. Combining this module with MS-TCN improves the performance.
However, there is still room for improvement, especially on large datasets with diverse action classes such as the Breakfast dataset~\cite{breakfast}.
Existing works perform with high frame-wise accuracy, but produce many false positive.
which results in over-segmentation error despite the multi-stage architecture and the smoothing loss.
Our experiments are conducted with comparison to these state-of-the-art methods to demonstrate the efficacy of our proposal.
Some works \cite{as_da, as_ssl_da} apply domain adaptation techniques to action segmentation,
but we do not compare these works as their setting is different from our work.

\textbf{Action Proposal Generation.}
Substantial research has been carried out in this domain, some of which is related to our approach.
Existing methods for action proposal generation can broadly be divided into two types of approaches: anchor-based approaches~\cite{ssad,cbr,turntap} and anchor-free approaches~\cite{tag,bsn,bmn}.
Anchor-based approaches define multi-scale anchors with even intervals as proposals and generate confidence scores for each proposal.
Anchor-free approaches first evaluate actionness or the likelihood of a frame being the start or end of an action, and then generate final predictions by leveraging these cues.
Inspired by these anchor-free approaches, we additionally use an action boundary regression network for the action segmentation task.
Our action boundary network regresses only boundary probabilities, regardless of it being a start or end of an action.
Using predicted action boundaries, our framework refines frame-wise predictions to improve the performance of action segmentation.


\section{Our Proposed Method}

In this section, we introduce our approach for action segmentation, ASRF.
Our framework decouples frame-wise action classification and action boundary regression.
Our proposed framework consists of a long-term feature extractor and two branches, an Action Segmentation Branch (ASB) and a Boundary Regression Branch (BRB), as in Figure~\ref{fig:network}.
A long-term feature extractor takes video features as input, expands the receptive field and captures long-term dependencies between action segments.
Then, both branches take the features as inputs and frame-level action predictions and action boundary probabilities as outputs. 
Let $X = [x_1, ... x_T ] \in \mathbb{R}^{T \times D}$ be the input to the ASRF, where T is the number of frames in a video and $D$ is the dimension of the feature. 
Given $X$, our goal is to classify frame-level action classes $C = [c_1, ..., c_T]$.
For each frame, we predict action boundaries $B = [b_1, ..., b_T]$, and use this to improve the classification. 

 

\subsection{Long-term Feature Extractor}
\label{extractor}
Given $X$, the goal of a long-term feature extractor is to capture long-term dependencies between action segments and extract rich features $X' \in \mathbb{R}^{T \times D'}$, where $D'$ is the dimension of the feature.
As a long-term feature extractor, we use a temporal convolutional network (TCN) with dilated residual layers proposed in \cite{mstcn}.
This architecture can convolve features with full temporal resolution and a large receptive field.
This enables the network to capture long-term dependencies between action segments and extract shared features for the following branches. 
Our feature extractor consists of 10 dilated residual layers with 64 filters, each of which are followed by a dropout layer with a dropout rate of 0.5.
The dilation rate is doubled at every residual convolution.

\subsection{Action Segmentation Branch}
\label{asb}
Given $X'$, the goal of the Action Segmentation Branch (ASB) is to predict frame-wise action classes $C$.
To predict action classes, we simply use a 1D convolutional layer followed by a softmax layer.
However this prediction contains some errors such as over-segmentation errors.
So we add a multi-stage architecture proposed in \cite{mstcn} after the output layer.
The first layer takes $X'$ as input and outputs the initial predictions,
and the subsequent stages refine the predictions made from previous stages.
This architecture facilitates capturing temporal dependencies and recognizes action segments, 
preventing over-segmentation errors.
Herein, each stage consists of a single temporal convolution with a kernel size of 1 and 64 filters,
10 dilated residual convolutions, and another temporal convolution
for reducing the feature dimension to the number of action classes. 
The parameters of each dilated convolutional layer are the same as in the long-term feature extractor.
We set the number of stages to 3 after the initial prediction layer as per \cite{mstcn}.

\subsection{Boundary Regression Branch}
\label{brb}
Although stacking several TCNs improves performance of action segmentation,
the predictions still contain over-segmentation errors.
To address this, we introduce a Boundary Regression Branch (BRB) in addition to the ASB.
Given $X'$, the BRB aims to regress the action boundary probabilities $P \in [0, 1]^T$ in a video,
which are used later for refining the action segmentation results from the ASB (see Section \ref{sec:refine}).
Action boundaries are defined as frames when an action starts and ends irrespective of action classes.
Unlike the methods which use a Hidden Markov model to determine the most probable sequence of actions~\cite{markov1,hmm},
the BRB is class-agnostic, which eliminates the need for modeling the probabilities from each class to every other.
Rather, the BRB only regresses the likelihood of general action boundaries.
Therefore, the BRB requires far less data for training and can improve the robustness in comparison with the class-aware methods.
We applies the the same structure as in the ASB to the BRB.
The stacked structure also allows the refinement of action boundary predictions within the branch.
In Section \ref{sec:num_stages}, we will explore the effect of the number of stages for action boundary regression.


\subsection{Refining Action Segmentation Results}
\label{sec:refine}
We describe how to refine the action segmentation results $C$ from the ASB using action boundary probabilities $P_b$ from the BRB.
First, we determine action boundaries $B \in \{0, 1\}^T$ from $P_b$.
We define $B$ as the frame-level prediction where $P_{b,t}$ scores a local maximum and is over a certain threshold $\theta_p$.
$B$ is our prediction for action boundaries, so we divide action segments based on these predictions.
Assuming that each segment contains a single action, 
we assign action classes to segments based on the action segmentation predictions from the ASB.
We make final predictions by majority voting on the action class within each segment.
Note that this refinement process is done only during inference.
In the experiments, the efficacy of this refinement strategy will be illuminated (see Section \ref{sec:impact_asrf}).

\subsection{Loss Function}
\label{sec:loss}
Our framework outputs both frame-wise action predictions and action boundaries.
Hence, our loss function is defined as:
\begin{equation}
    \mathcal{L}=\mathcal{L}_{asb} + \lambda \mathcal{L}_{brb}
\end{equation}
where $\mathcal{L}_{asb}$ and $\mathcal{L}_{brb}$ are loss functions for the ASB and the BRB respectively
and $\lambda$ is the weight of the $\mathcal{L}_{brb}$.
In our work, we set $\lambda$ to 0.2 for GTEA and 0.1 for 50 Salads and the Breakfast dataset.
In the following sections, we describe loss functions for each branch. 

\subsubsection{Loss Function for ASB.}
\label{sec:loss_asb}
Existing works usually adopt cross entropy as a classification loss.
\begin{equation}
\mathcal{L}_{ce}=\frac{1}{T} \sum_{t}-\log \left(y_{t, c}\right)
\end{equation} \noindent
where $y_{t,c}$ is the action probability for class $c$ at time $t$.
However, this approach cannot penalize over-segmentation errors because there is no constraint for temporal transition of probabilities. To overcome this, the authors of~\cite{mstcn} additionally use the Truncated Mean Squared Error (TMSE).
\begin{align}
\mathcal{L}_{TMSE}&=\frac{1}{T N} \sum_{t, c} \tilde{\Delta}_{t, c}^{2} \\
\tilde{\Delta}_{t, c}&=\left\{\begin{array}{ll}{\Delta_{t, c}} & {: \Delta_{t, c} \leq \tau} \\ {\tau} & {: \textit {otherwise}}\end{array}\right.\\
\Delta_{t, c}&=\left|\log y_{t, c}-\log y_{t-1, c}\right|
\end{align}
\noindent
where $T$ is the length of a video, $N$ is the number of classes and $\tau$ is a threshold for the transition of probabilities.

Herein, in addition to these two loss functions, we validate two other loss functions.
First, we simply impose a class weight for the cross entropy loss $L_{ce, cw}$).
The frequency of different action segments differs for each action class, leading to an imbalance during training.
For weighting, we use median frequency balancing~\cite{median},
where the weight to each class is calculated by dividing the median of class frequencies by each class frequency.
In our experiments, we also compare this weighting method with Focal Loss \cite{focal} (see Section \ref{sec:combination_loss}).

Next, we introduce Gaussian Similarity-weighted TMSE (GS-TMSE) as a loss function, which improves upon TMSE.
TMSE penalizes all frames in a video to smooth the transition of action probabilities between frames.
However, this results in penalizing the frames where actions actually transition.
To address this problem, we apply the Gaussian kernel to TMSE as follows:
\begin{equation}
\mathcal{L}_{GS-TMSE}=\frac{1}{T N} \sum_{t, c} \exp \left(-\frac{\left\|\mathbf{x}_{t}-\mathbf{x}_{t-1}\right\|^{2}}{2 \sigma^{2}}\right) \tilde{\Delta}_{t, c}^{2}
\end{equation}
\noindent
where $\mathbf{x_t}$ is an index of similarity for frame $t$ and $\sigma$ denotes variance.
Because of the Gaussian kernel based on the similarity of frames, this function penalizes adjacent frames with large differences with a smaller weight.
Here, we use the frame-level input feature for an index of similarity and set $\sigma$ to 1.0.
We also set $\tau$ in TMSE and GS-TMSE as 4, following~\cite{mstcn}.

The loss function for each prediction in the ASB is defined:
\begin{equation}
\label{loss_asb}
    \mathcal{L}_{as}= \mathcal{L}_{ce} + \mathcal{L}_{GS-TMSE} 
\end{equation}
Then we average the losses for each prediciton in the ASB as follows:
\begin{equation}
\label{loss_asb}
    \mathcal{L}_{asb}= \frac{1}{{N}_{as}} \sum_{i} \mathcal{L}_{as, i}
\end{equation}
where $N_{as}$ is the number of predictions in the ASB ($N_{as}$ = 4 in our framework).

In the experiments, we compare various loss functions and their combinations for action segmentation as well as Equation~\ref{loss_asb}. When combining them, we simply add each loss function except TMSE. We multiply $L_{TMSE}$ by 0.15 and then add the other loss functions as per~\cite{mstcn}.

\subsubsection{Loss Function for BRB.}
\label{sec:loss_brb}
We use a binary logistic regression loss function for the action boundary regression:
\begin{equation}
\mathcal{L}_{bl}
=\frac{1}{T} \sum_{t=1}^{T}\left(w_{p} y_{t} \cdot \log p_{t} +\left(1-y_{t}\right) \cdot \log \left(1-p_{t}\right)\right)
\end{equation}
\noindent
where $y_t$ and $p_t$ are the ground truth and the action boundary probability for frame $t$, respectively. 
We weight positive samples by $w_p$ since the number of frames that are action boundaries is much smaller than that of the others. 
We calculate the ratio of positive data points over the whole training data and use the reciprocal of this as the weight.
As in the ASB, we average the losses for each boundary prediction in the BRB as follow:
\begin{equation}
\label{loss_asb}
    \mathcal{L}_{brb}= \frac{1}{{N}_{br}} \sum_{i} \mathcal{L}_{bl, i}
\end{equation}
where $N_{br}$ is the number of predictions in the BRB ($N_{br}$ is also 4 in our framework).


\section{Experiments}

\textbf{Datasets.}
For our evaluation, we use three challenging datasets: 50 Salads~\cite{50salads}, Georgia Tech Egocentric
Activities (GTEA)~\cite{gtea}, and the Breakfast dataset~\cite{breakfast}.
The 50 Salads dataset contains 50 videos in which 25 people in total are preparing two kinds of mixed salads.
The videos, each consisting of 9000 to 18000 RGB frames, depth maps and accelerometer data, are annotated with 17 action classes every frame. 
The GTEA dataset contains 28 videos of 7 types of daily activities, each performed by 4 different subjects.
Each video is recorded egocentrically, from a camera mounted on a subject's head.
The Breakfast dataset contains over 77 hours of videos, where 10 classes of actions are performed by 52 individuals in 18 different kitchens.

As per~\cite{mstcn}, for all datasets, we use spatiotemporal features extracted by I3D~\cite{CarreiraCVPR2017} as input to both ASB and BRB.
Following~\cite{mstcn}, we downsample the videos in the 50 Salads dataset from 30 to 15 fps to ensure consistency between the datasets.
For evaluation, we use 5-fold cross-validation on 50 Salads dataset and 4-fold cross-validation on the others as in \cite{50salads,breakfast,mstcn}.

\begin{figure}[t!]
\begin{center}
\includegraphics[width=70mm]{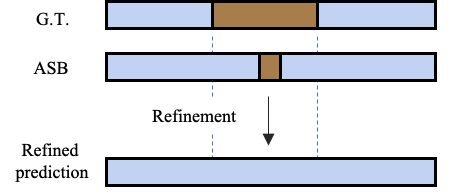}
\end{center}
\caption{An example where the prediction refined the by ground truth boundaries is worse than that of the ASB. 
The color bands differentiate action classes, and the horizontal direction denotes time.
Our framework decides action classes of each segment by majority voting. Therefore segments can be reassigned as the wrong class.}
\label{fig:oracle}
\vspace{-5pt}
\end{figure}

\begin{figure*}[t!]
\begin{center}
\includegraphics[width=170mm]{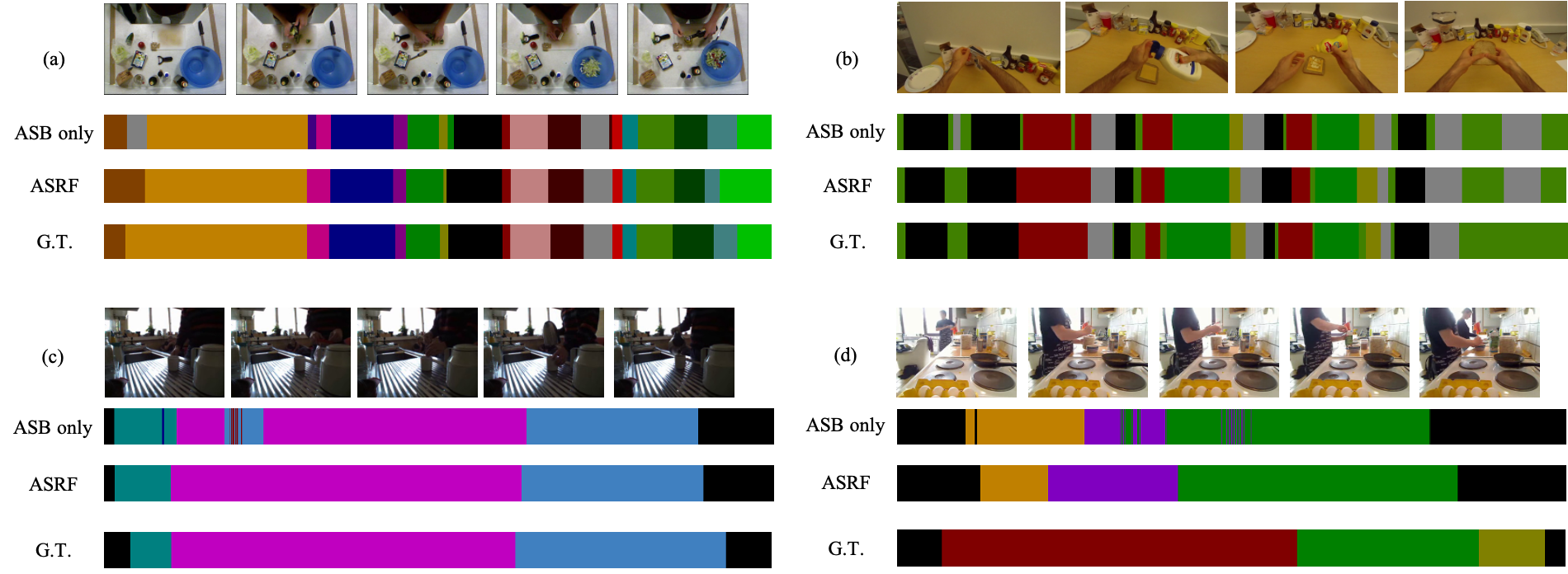}
\end{center}
\caption{Qualitative results using (a) 50 Salads, (b) GTEA and (c) Breakfast dataset in comparison with predictions before refinement. (d) Failure example on Breakfast dataset}
\label{fig:qualitative}
\vspace{-5pt}
\end{figure*}

\begin{table*}[t!]
\small
\centering
\begin{tabular}{c | ccccc |ccccc }
\hline
Dataset             & \multicolumn{5}{c}{50 Salads}                        & \multicolumn{5}{c}{GTEA}                            \\ \hline
Method               & \multicolumn{3}{c}{F1@\{10, 25, 50\}} & Edit & Acc  & \multicolumn{3}{c}{F1@\{10, 25, 50\}} & Edit  & Acc  \\ \hline
Bi-LSTM \cite{bi-lstm2}       & 62.6   & 58.3   & 47.0   & 55.6   & 55.7                              & 66.5   & 59.0   & 43.6   & -      & 55.5            \\ 
ED-TCN \cite{tcn}             & 68.0   & 63.9   & 52.6   & 59.8   & 64.7                              & 72.2   & 69.3   & 56.0   & -      & 64.0            \\ 
TDRN \cite{deform}            & 72.9   & 68.5   & 57.2   & 66.0   & 68.1                              & 79.2   & 74.4   & 62.7   & 74.1   & 70.1            \\ 
MS-TCN \cite{tcn}             & 76.3   & 74.0   & 64.5   & 67.9   & 80.7                              & 85.8   & 83.4   & 69.8   & 79.0   & 76.3            \\ 
MS-TCN + BPGaussian \cite{bpgaussian}  & 78.4   & 75.8   & 66.7   & 71.0   & 80.6                     & 86.7   & 84.3   & 72.7   & 77.2   & \textbf{82.3}  \\
MS-TCN + GTRM \cite{as_graph} & 75.4   & 72.8   & 63.9   & 67.5   & 82.6                              & -      & -      & -      & -      & -               \\ \hline
ASRF                          & \textbf{84.9} & \textbf{83.5} & \textbf{77.3} & \textbf{79.3} & \textbf{84.5}    & \textbf{89.4} & \textbf{87.8} & \textbf{79.8} & \textbf{83.7} & 77.3  \\
\rowcolor[gray]{0.90}
Improvement                   & \textbf{+6.5}    & \textbf{+5.1}    & \textbf{+10.6}   & \textbf{+8.3}    & \textbf{+1.9}   & \textbf{+2.7}    & \textbf{+3.5}    & \textbf{+7.1}    & \textbf{+4.7}    & -5.0 \\ \hline
\end{tabular}
\caption{Comparing our proposed method with existing methods on 50 Salads and GTEA.}
\label{tab:50salads gtea}
\end{table*}

\textbf{Evaluation Metrics for Action Segmentation.}
The following three metrics are used for action segmentation as in~\cite{icra,markov2,tcn}: frame-wise accuracy (Acc), segmental edit distance (Edit), and segmental F1 score with overlapping threshold $k\%$ (F1@k). 
Although frame-wise accuracy is commonly used as a metric for action segmentation, this measure is not sensitive to over-segmentation errors.
Therefore, as well as frame-wise accuracy, we also use segmental edit distance~\cite{icra,markov2} and segmental F1 score~\cite{tcn} because both of the latter penalize over-segmentation errors.

The segmental edit distance, $S_{edit}(G, P)$ is a metric for measuring the difference between ground truth segments $G = \{G_1, ..., G_M\}$ and predicted segments $P = \{P_1, ..., P_N\}$.
This metric is calculated using the Levenshtein Distance\cite{levenshtein} between hypotheses and ground truths.
For the sake of clarity, we report $(1 - S_{edit}(G, P) / max(M, N)) \times 100$ as segmental distance.

The segmental F1 score is averaged per class, in which a prediction is classified as correct if the temporal Intersection over Union (IoU) is larger than a certain threshold.
This metric is invariant with respect to temporal shifts in predictions emanating from the ambiguity of the action boundary or human annotation noise.

\textbf{Evaluation Metrics for Boundary Regression.}
The action boundary F1 score is used as an evaluation metric for the action boundary regression.
We define the action boundary F1 score referencing the boundary F1 score used for semantic segmentation~\cite{boundaryf1}.
Let $B_{gt} \in \{0, 1\}^T$ denote if each frame is a boundary or not and $P_{b} \in [0, 1]^T$ denote the predicted boundary probability map.
We define $B_{pred} \in \{0, 1\}^T$ as the frame-level prediction where $P_{b, t}$ is both over a threshold $\theta_{p}$ and is a local maximum.
The precision and the recall for action boundaries are defined as:
\begin{eqnarray}
    Precision=\frac{1}{\left|B_{pred}\right|} \sum_{x \in B_{pred}}I\left[d\left(x, B_{gt}\right)<\theta_{b}\right], \\
    Recall=\frac{1}{\left|B_{gt}\right|} \sum_{x \in B_{gt}}I\left[d\left(x, B_{pred}\right)<\theta_{b}\right],
\end{eqnarray}
where $I[\cdot]$ denotes the indicator function, $d(\cdot)$ is the L1 distance for temporal span and $\theta_{b}$ is a threshold over timestamp.
We set $\theta_{b}$ as 5 and $\theta_p$ as 0.5 in all experiments.
Then the boundary F1 metric is defined as $BF=\frac{2 \times Precision \times Recall}{Precision + Recall}$.

In addition, assuming that we have an oracle ASB which outputs the ground truth frame-wise labels,
we use the BRB and refine the output of this oracle ASB.
Using a single BRB, this is the upper bound of the achievable score.
Therefore, when comparing multiple BRBs, this upper bound can be used to evaluate the efficacy. 
Note that we do not assume an oracle BRB for refining action segmentation results from the ASB, as upper bounds are not achievable even with an oracle BRB (See Figure~\ref{fig:oracle}).

\textbf{Learning Scheme.}
We train the entire framework using the Adam optimizer with a learning rate of 0.0005 and batch size of 1 as per~\cite{mstcn}.
We also find the optimal number of epochs with nested cross-validation.
During inference, action segmentation results from the ASB are refined using predicted action boundaries from the BRB. 


\subsection{Comparing ASRF with the state-of-the-art}
\label{sec:sota}
We compare our proposed framework with existing methods on three challenging datasets: 50 Salads, Georgia Tech Egocentric Activities (GTEA) and the Breakfast dataset. 
Table \ref{tab:50salads gtea} shows the results for the first two datasets.

Therein, our ASRF is superior to the state-of-the-art in terms of segmental edit distance and segmental F1 score with competitive frame-wise accuracy on each dataset,
having up to 8.3\% improvement for segmental edit distance, and up to 10.6\% improvement for the segmental F1 score on 50 Salads and GTEA.
As per Table \ref{tab:breakfast}, our framework also outperforms the existing methods by a large margin on the Breakfast dataset with respect to all evaluation metrics.

We find that our framework offers better results for stricter overlap thresholds when calculating the segmental F1 score.
On GTEA, our framework and MS-TCN with bilinear pooling \cite{bpgaussian} perform similarly in terms of F1@10.
However, our framework outperforms \cite{bpgaussian} on F1@50 by 7.1\%.
This shows that our framework is capable of recognizing action segments which overlap markedly with the ground truth segments.

\begin{figure*}[h]
    \small
    \begin{center}
    \includegraphics[width=171mm]{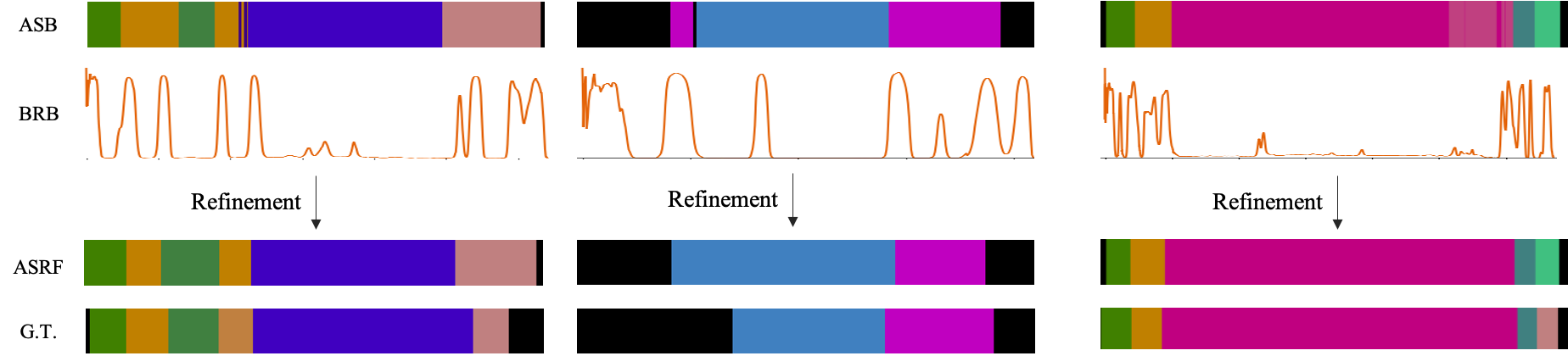}
    \end{center}
    \caption{Examples of refinement process on the Breakfast dataset. The first row shows predictions from the ASB and the second row shows predicted boundary probabilities.
    Combining them, the ASRF outputs final predictions (the third row).}
    \label{fig:refine}
    \vspace{-5pt}
\end{figure*}

In the case of GTEA, the frame-wise accuracy for our framework is inferior to that of \cite{bpgaussian}.
We hypothesize that our framework predicted action boundaries off by some margin, therefore affecting the frame-wise accuracy.
The rise in the segmental metrics support our hypothesis, as these errors are not accounted for here.

Qualitative results are presented in Figure~\ref{fig:qualitative}. 
Predictions before refinement have some over-segmentation errors, but our framework can reduce these using action boundaries (Figure~\ref{fig:qualitative} (a), (b), (c)).
Figure~\ref{fig:qualitative} (d) shows a failure case of our framework.
The BRB has limits in the sense that it cannot reassign completely incorrect segments
of inferred action classes by the ASB.

\subsection{Effect of our refining paradigm}
\label{sec:impact_asrf}
In this section, we show the effect of refinement on the action segmentation metrics using action boundaries predicted by the BRB.
In addition to the refinement with the BRB, we evaluate three other postprocessing methods:
\textit{i)} \textbf{Relabeling.} relabeling actions of segments shorter than a certain temporal span $\theta_t$ with the action of the previous segment,
\textit{ii)} \textbf{Smoothing.} smoothing action probabilities using the 1D Gaussian filter with a kernel size $K$, and
\textit{iii)} \textbf{Similarity.} refinement with predicted action boundaries based on frame-level similarity.
We measure the similarity of frames using frame-level features as in Section \ref{sec:loss_asb},
and decide action boundary positions based on where the similarity is a local minimum. 

Table \ref{tab:BRN} shows the action segmentation results of the ASB and those after refinement on the 50 Salads dataset.
As observed, our refinement method with action boundaries improves action segmentation results by over 8.8\% for every segmental metric with comparable frame-wise accuracy.
Especially the stricter we set the IOU threshold for the segmental F1 score, the better the ASRF scores.
This shows that the ASRF not only prevents over-segmentation errors but also generates predictions that highly overlap with the ground truth.
In addition, our method is superior to other ways of postprocessing in terms of every metric.
Smoothing has less impact on the metrics than our ASRF, and Similarity has a negative effect.
Although Relabeling is better than the other two methods, this method is highly dependent on the hyperparameter $\theta_t$.
With a large $\theta_t$, the method can reduce over-segmentation errors, but drops small action segments.
On the other hand, our framework can detect action boundaries and capture action segments irrespective of temporal length, which results in better performance.

Figure~\ref{fig:refine} shows our refinement process on the Breakfast dataset.
The BRB outputs some false positives, but they are ignored when the ASB captures action segments without over-segmentation errors (the middle of Figure~\ref{fig:refine}).
This shows that our framework not only refines frame-level action predictions with boundaries,
but selects reasonable action boundaries during refinement as well. 
Therefore, the ASB and the BRB are mutually supportive.

\begin{table}[t!]
\small
\centering
\begin{tabular}{cccccc}
\hline
Breakfast          & \multicolumn{3}{c}{F1@\{10, 25, 50\}} & Edit   & Acc   \\ \hline
ED-TCN \cite{tcn} *           & -          & -          & -          & -       & 43.3  \\ 
HTK \cite{htk}                & -          & -          & -          & -       & 50.7  \\ 
TCFPN \cite{softboundary}     & -          & -          & -          & -       & 52.0  \\ 
HTK(64) \cite{markov1}        & -          & -          & -          & -       & 56.3  \\ 
GRU \cite{softboundary} *     & -          & -          & -          & -       & 60.6  \\ 
MS-TCN \cite{mstcn}           & 58.2       & 52.9       & 40.8       & 61.4    & 65.1  \\ 
MS-TCN+GTRM~\cite{as_graph}   & 57.5       & 54.0       & 43.3       & 58.7    & 65.0     \\\hline
ASRF                          & \textbf{74.3}   & \textbf{68.9}   & \textbf{56.1}  &    \textbf{72.4} & \textbf{67.6}   \\ 
\rowcolor[gray]{0.90}
Improvement                   & \textbf{+16.1}       & \textbf{+14.9}    & \textbf{+12.8}   & \textbf{+13.7}  & \textbf{+2.5} \\ \hline
\end{tabular}
\caption{Comparing our proposed method with existing methods on the Breakfast dataset. * reported results by the author of \cite{softboundary}.}
\label{tab:breakfast}
\vspace{-10pt}
\end{table}

\begin{table}[t]
\small
\centering
\begin{tabular}{cccccc}
\hline
                              & \multicolumn{3}{c}{F1@\{10, 20, 50\}} & Edit & Acc \\ \hline
w/o post-processing           & 76.1          & 74.5          & 66.7          & 68.5          & 82.6 \\
Relabeling ($\theta_t = 5$)   & 81.5          & 79.7          & 71.6          & 74.8          & 82.6 \\
Relabeling ($\theta_t = 15$)  & 82.4          & 80.6          & 72.5          & 76.2          & 82.7 \\
Smoothing ($K = 5$)           & 80.7          & 78.9          & 70.8          & 73.5          & 82.6 \\
Smoothing ($K = 15$)          & 80.8          & 79.0          & 70.9          & 73.6          & 82.6 \\
Similarity                    & 39.8          & 30.9          & 18.1          & 32.8          & 40.0\\ \hline
ASRF                          & \textbf{84.9} & \textbf{83.5} & \textbf{77.3} & \textbf{79.3} & \textbf{84.5}  \\ \hline
\end{tabular}
\caption{Effect of refinement strategy using action boundaries from the BRB on the 50 Salads dataset.}
\label{tab:BRN}
\vspace{-3pt}
\end{table}

\begin{table}[t]
\small
\centering
\begin{tabular}{cccccc}
\hline
                                  & \multicolumn{3}{c}{F1@\{10, 20, 50\}} & Edit & Acc \\ \hline
Segment classifier                & 51.8   & 49.0   & 40.0   & 42.8   & 58.5   \\
ASRF                              & \textbf{84.9} & \textbf{83.5} & \textbf{77.3} & \textbf{79.3} & \textbf{84.5}   \\ \hline
\end{tabular}
\caption{Comparing our proposed method with the segment-level classification method on the 50 Salads dataset.}
\label{tab:seg_lev_cls}
\vspace{-5pt}
\end{table}

\begin{table*}[t!]
\small
\centering
\begin{tabular}{cccc|ccccc|ccccc}
\hline
                    & \multicolumn{3}{c}{Boundary Regression}  & \multicolumn{5}{c}{Action Segmentation}             & \multicolumn{5}{c}{Action Segmentation (Oracle)}\\ \hline
                    & Precision      & Recall      & F1 Score  & \multicolumn{3}{c}{F1@\{10, 20, 50\}} & Edit & Acc  & \multicolumn{3}{c}{F1@\{10, 20, 50\}} & Edit & Acc\\ \hline
No stage            & 18.8  & 76.4   & 29.9 & 82.0         & 80.4          & 74.4          & 76.7          & 82.7           & 86.8   & 86.2   & 83.5   & 82.2   & 88.5 \\
1 stage            & 34.8  & 69.9   & 46.4 & 82.5          & 81.1          & 73.1          & 76.8          & 80.9           & 86.0   & 85.4   & 81.7   & 81.4   & 87.2 \\
2 stages            & 38.7  & 66.0   & 48.7 & 83.3          & 82.3          & 76.4          & 78.3          & 82.7           & 86.2   & 85.8   & 84.1   & 82.7   & 88.6 \\
3 stages            & 37.3  & 63.2   & 46.8 & \textbf{84.9} & \textbf{83.5} & \textbf{77.3} & \textbf{79.3} & \textbf{84.5}  & 86.9   & 86.8   & 84.5   & 83.2   & 89.3 \\
4 stages            & 37.4  & 63.0   & 46.9 & 84.4          & 82.9          & 75.7          & 78.1          & 82.7           & 86.5   & 86.0   & 83.8   & 82.5   & 87.8 \\ \hline
\end{tabular}
\caption{Effect of the number of stages for the BRB after the initial prediction layer on the 50 Salads dataset.}
\label{tab:num_stages}
\vspace{-5pt}
\end{table*}

\begin{table*}[t!]
\small
\centering
\begin{tabular}{cccc|ccccc}
\hline
          & \multicolumn{3}{c}{Boundary Regression} & \multicolumn{5}{c}{Action Segmentation}                           \\ \hline
$\theta_p$ & Precision     & Recall    & F1 Score    & \multicolumn{3}{c}{F1@\{10, 20, 50\}} & Edit     & Acc            \\ \hline
0.1        & 36.5      & 63.7    & 46.3       & 84.6          & 83.3          & 76.7          & 79.0          & 84.4             \\
0.3        & 37.0      & 63.4    & 46.7       & 84.7          & 83.4          & 76.8          & 78.9          & 84.4                  \\
0.5        & 37.3      & 63.2    & 46.8       & \textbf{84.9} & \textbf{83.5} & \textbf{77.3} & \textbf{79.3} & \textbf{84.5}  \\
0.7        & 37.6      & 63.1    & 47.1       & 84.5          & 83.2          & 77.3          & 78.8          & 83.9                  \\
0.9        & 37.6      & 63.1    & 47.1       & 84.5          & 83.2          & 77.3          & 78.8          & 83.9                  \\ \hline
\end{tabular}
\caption{Comparing the effect of $\theta_p$ for action boundary decision}
\label{tab:threshold}
\vspace{-5pt}
\end{table*}

\begin{table}[t!]
\small
\centering
\begin{tabular}{c|ccccc}
\hline
Loss Function &  \multicolumn{3}{c}{F1@\{10, 20, 50\}} & Edit & Acc \\ \hline
$\mathcal{L}_{ce}$ + $\mathcal{L}_{TMSE}$              & 79.7          & 77.7          & 69.1          & 73.4          & 79.8 \\
$\mathcal{L}_{ce}$ + $\mathcal{L}_{GS-TMSE}$           & 80.9          & 79.4          & 72.7          & 74.6          & 81.6 \\
$\mathcal{L}_{focal}$ + $\mathcal{L}_{TMSE}$              & 77.3          & 75.8          & 67.1          & 71.3          & 78.0\\
$\mathcal{L}_{focal}$ + $\mathcal{L}_{GS-TMSE}$           & 78.3          & 76.0          & 66.3          & 71.0          & 78.0\\
$\mathcal{L}_{ce, cw}$ + $\mathcal{L}_{TMSE}$    & 83.3          & 81.8          & 75.1          & 77.0          & 81.8\\
$\mathcal{L}_{ce, cw}$ + $\mathcal{L}_{GS-TMSE}$ & \textbf{84.9} & \textbf{83.5} & \textbf{77.3} & \textbf{79.3} & \textbf{84.5}   \\ \hline
\end{tabular}
\caption{Comparing combinations of loss functions on the 50 Salads dataset.}
\label{tab:loss}
\end{table}

\subsection{Comparison with segment-level classifier}
\label{sec:seg_lev_cls}
Our framework classifies actions by the frame level before predicting action boundaries, then reassigning action classes for each predicted action segment.
Another variant is to use a TCN to classify each predicted action segment.
Therefore, we compare our framework with a combination of a boundary regression model and segment-level classifier.
We use a single-stage TCN with 10 dilated convolutions and two convolutions as a segment-level classifier.
Note that we add global average pooling before the last convolutional layer of the single-stage TCN to aggregate temporal features.
Otherwise, for both the action boundary regression and segment-level classification, we use the same networks as our method.
As seen in Table \ref{tab:seg_lev_cls}, our framework outperforms this variant by a large margin.
This shows the importance of capturing long-range dependencies between action segments in the context of action segmentation.

\subsection{Ablation Study}
\subsubsection{Effect of the number of stages}
\label{sec:num_stages}
\vspace{-2pt}
We use a multi-stage architecture for BRB as well as ASB, which must be validated.
Therefore we train a variety of multi-stage networks and evaluate their performances.
Each network stage has 10 dilated convolutions and two convolutions like our proposed BRB.
As can be observed in Table \ref{tab:num_stages}, the three-stage architecture after the initial prediction layer outperforms others in terms of action segmentation.
This shows that stacking TCNs helps to regress boundaries as well.
However, using four stages on the 50 Salads dataset does not improve the predictions, which is thought to be the result of overfitting due to the size of the 50 Salads dataset.

We found that the precision of the BRB is low, but overdetected boundaries can still help the refinement of frame-wise action predictions. 
The reason of the low precision is that it is difficult to predict action boundaries precisely because of the ambiguity of human annotations and the action boundary itself.
In addition, the $\theta_b$ for calculating precision is only 5 frames, which is a strict threshold.
Also, we highlight that the low precision of the BRB does not have a negative impact.
In our framework, the ASB and the BRB are complementary, so even if nonexistent action boundaries are predicted, some of them disappear when combining outputs from the ASB.
In addition, segmental edit distance and segmental F1 score with IoU thresholds tolerate the shifts of action starting and ending points.
Results show that there is little to no correlation between boundary scores and action segmentation scores.
This is because segmental metrics are tolerant to shifts in action boundary predictions.
Therefore, oracle experiments are selected for the evaluation of boundary regression in our framework.

\vspace{-2pt}
\subsubsection{Impact of $\theta_p$ for boundary regression}
As described in Section \ref{sec:loss_brb}, we set the threshold $\theta_p = 0.5$ for deciding action boundaries from outputs predicted by BRB.
Table \ref{tab:threshold} shows the impact of $\theta_p$ for boundary regression.
Note that we do not use an oracle ASB in this experiment, because an overdetecting BRB would obtain perfect results, therefore setting $\theta_p$ to a minimal value would be best performing.
As in Table \ref{tab:threshold}, our framework is not particularly sensitive to $\theta_p$.
For our framework, $\theta_p = 0.5$ showed the best balance.

\vspace{-2pt}
\subsubsection{Comparing loss functions for the ASB}
\label{sec:combination_loss}
Table \ref{tab:loss} compares the results of each combination of loss functions.
Our proposed smoothing loss $\mathcal{L}_{GS-TMSE}$ improves the segmental metrics by up to 3.6\%.
This shows that $\mathcal{L}_{GS-TMSE}$ can smooth the transition of action probabilities without penalizing frames where actions actually transition, which results in alleviating over-segmentation errors.
In addition, imposing class weight to the cross also improves all the metrics
while $\mathcal{L}_{focal}$ has a negative influence.
Though tuning hyperparameters in $\mathcal{L}_{focal}$ may still help to improve performance,
it is costly to do so for every dataset.
Our combination of loss functions enable better performance in comparison with loss functions used in existing works. 


\vspace{-2pt}
\section{Conclusions}
We proposed an effective framework for the action segmentation task.
In addition to an action segmentation network, we use an action boundary regression network for refining action segmentation results with predicted action boundaries.
We also compared and evaluated various loss functions and their combinations for action segmentation.
Through experiments, it was confirmed that our framework is capable of recognizing action segments and alleviating over-segmentation errors.
Our framework outperforms state-of-the-art methods on the three challenging datasets, especially on the Breakfast dataset by a large margin,
which contains a larger number of videos and action classes than the others.

\section*{Acknowledgement}
Computational resource of AI Bridging Cloud Infrastructure (ABCI) provided by National Institute of Advanced Industrial Science and Technology (AIST) was used.

\bibliographystyle{ieee_fullname}
\bibliography{strings,refs}


\end{document}